\documentclass[10pt,conference,a4paper]{IEEEtran}
\usepackage[left=1.34cm,right=1.34cm,top=1.65cm,bottom=4.05cm]{geometry}
\usepackage{acronym}
\acrodef{3GPP}{3rd generation partnership project}
\acrodef{3D}{three-dimensional}
\acrodef{2D}{two-dimensional}
\acrodef{5G}{fifth generation}
\acrodef{6G}{the sixth generation}
\acrodef{AI}{artificial intelligence}
 \acrodef{AGV}{automatic guided vehicle}
 \acrodef{ADMM}{alternating direction method of multipliers}
\acrodef{BS}{base station}
\acrodef{CCO}{coverage and capacity optimization}
\acrodef{CDF}{cumulative distribution function}
\acrodef{CIO}{cell individual offset}
\acrodef{CTDE}{centralized training with decentralized execution}
\acrodef{CSI}{channel state information}
\acrodef{CQI}{channel quality indicator}
\acrodef{CNN}{convolutional neural network}
\acrodef{CSP}{communications service provider}
\acrodef{CV}{computer vision}
\acrodef{CA}{cell-level agent}
\acrodef{CPA}{cell pair-level agent}
\acrodef{CDRL}{compositional deep reinforcement learning}
\acrodef{CPDM}{compositional predictive decision-making}
\acrodef{MACDRL}{multi-agent compositional deep reinforcement learning}
\acrodef{MACPDM}{multi-agent compositional predictive decision making}
\acrodef{DL}{deep learning}
\acrodef{DNN}{deep neural network}
\acrodef{DQN}{deep Q-network}
\acrodef{DRL}{deep reinforcement learning}
\acrodef{DDPG}{Deep Deterministic Policy Gradient}
\acrodef{XR}{extended reality}
\acrodef{FDD}{frequency division duplex}
\acrodef{FDMA}{frequency division multiple access}
\acrodef{GP}{Gaussian process}
\acrodef{GD}{Gradient Descent}
\acrodef{GUI}{graphical user interface}
\acrodef{HetNet}{heterogeneous network}
\acrodef{HO}{handover}
\acrodef{IoT}{internet of things}
\acrodef{IDLA}{integrated deep learning and Lagrangian method}
\acrodef{KL}{Kullback-Leibler}
\acrodef{KKT}{Karush–Kuhn–Tucker}
\acrodef{KPI}{key performance indicator}
\acrodef{LTE}{long term evolution}
\acrodef{M2M}{machine-to-machine}
\acrodef{MAC}{media access control}
\acrodef{MAE}{mean absolute error}
\acrodef{MAPE}{mean absolute percentage error}
\acrodef{MARL}{multi-agent reinforcement learning}
\acrodef{MADRL}{multi-agent deep reinforcement learning}
\acrodef{MDP}{Markov decision process}
\acrodef{ML}{machine learning}
\acrodef{MMDP}{multi-agent Markov Decision Process}
\acrodef{MIMO}{multiple-input and multiple-output}
\acrodef{mmWave}{millimeter wave}
\acrodef{MLP}{multi-layer perceptron}
\acrodef{MLB}{mobility load balancing}
\acrodef{MILP}{mixed integer linear programming}
\acrodef{MRO}{mobility robustness optimization}
\acrodef{NSM}{Network Slicing Management}
\acrodef{NLP}{natural language processing}
\acrodef{OAM}[O\&M]{Network Operation and Maintenance}
\acrodef{O-RAN}{Open Radio Access Network}
\acrodef{OFDM}{orthogonal frequency division multiplexing}
\acrodef{PASA}{production-aware slicing resource allocation}
\acrodef{PDF}{probability density function}
\acrodef{PF}{proportional fairness}
\acrodef{PPO}{proximal policy optimization}
\acrodef{PHY}{physical layer}
\acrodef{PSD}{power spectral density}
\acrodef{PRB}{physical resource block}
\acrodef{PRBs}{physical resource blocks}
\acrodef{QoE}{quality of experience}
\acrodef{QoS}{quality of service}
\acrodef{RAN}{radio access network}
\acrodef{RB}{resource block}
\acrodef{RL}{reinforcement learning}
\acrodef{RLF}{radio link failure}
\acrodef{RR}{round robin}
\acrodef{RRM}{radio resource management}
\acrodef{RSRP}{reference signal received power}
\acrodef{RU}{resource unit}
\acrodef{RX}{receiver}
\acrodef{SDN}{software defined network}
\acrodef{SNR}{signal-to-noise ratio}
\acrodef{SINR}{signal-to-interference-plus-noise ratio}
\acrodef{SIR}{signal-to-interference ratio}
\acrodef{SLA}{service level agreement}
\acrodef{SON}{Self-organizing networks}
\acrodef{SVM}{support vector machine}
\acrodef{TCP}{transmission control protocol}
\acrodef{TL}{Transfer learning}
\acrodef{TL-CPDM}{Transfer learning-enhanced CPDM}
\acrodef{TDD}{time division duplex}
\acrodef{TD}{temporal difference}
\acrodef{TD3}{Twin Delayed Deep Deterministic policy gradient}
\acrodef{TDMA}{time division multiple access}
\acrodef{TTI}{transmission time interval}
\acrodef{TTT}{time-to-trigger}
\acrodef{TX}{transmitter}
\acrodef{UDP}{user datagram protocol}
\acrodef{UE}{user equipment}
\acrodef{UL}{uplink}
\acrodef{V2X}{vehicle-to-everything}
\acrodef{WLAN}{wireless local area network}
\usepackage{acronym}
\usepackage{xspace}
\usepackage{verbatim}
\usepackage[usenames]{color}
\usepackage{array}
\usepackage{multicol}
\usepackage{algorithm}
\usepackage{algpseudocode}
\usepackage{amsmath,amsfonts,amssymb,amsthm}
\usepackage{graphicx}
\usepackage{subcaption}
\usepackage{float}
\usepackage{epstopdf}
\usepackage{cite}
\usepackage{url}

\usepackage[english]{babel}
\usepackage{enumitem} 
\usepackage[font=small,skip=2pt]{caption}
\usepackage[utf8]{inputenc}
\usepackage[mathscr]{euscript}
\usepackage{textcomp}
\usepackage{xcolor}
\usepackage{bbm}

\DeclareMathAlphabet\mathbfcal{OMS}{cmsy}{b}{n}

\newtheorem{Rem}{Remark}
\usepackage[english]{babel}
\usepackage{amsthm}
\usepackage{amssymb}
\usepackage{bbm}

\DeclareRobustCommand\optionalsec[1]{%
  \ifnum\pdfstrcmp{#1}{\thesection}=0\else#1.\fi
}
\usepackage{multirow}
\DeclareMathOperator*\argmax{arg \, max}		% arg max
\newcommand{\field}[1]{\mathbb{#1}}

\newcommand{\set}[1]{\mathcal{#1}}

\newcommand{\R}{{\field{R}}}   
\newcommand{\Ex}{{\field{E}}}
\newcommand{\NN}{{\field{N}}}                 

\newcommand{\1}{{\mathbbm{1}}}  

\newcommand{\ve}[1]{\boldsymbol{\mathbf{#1}}} % define all vector with bold lowercase letters

\newcommand{\vs}{\ve{s}}

\newcommand{\vp}{\ve{p}}

\newcommand{\vq}{\ve{q}}

\newcommand{\vg}{\ve{g}}

\newcommand{\va}{\ve{a}}
\newcommand{\vba}{\bar{\ve{a}}}
\newcommand{\vbs}{\bar{\ve{s}}}
\newcommand{\br}{\bar{r}}

\newcommand{\vh}{\ve{h}}
\newcommand{\vpsi}{\ve{\psi}}
\newcommand{\vrho}{\ve{\rho}}
\newcommand{\vphi}{\ve{\phi}}
\newcommand{\vthe}{\ve{\theta}}

\newcommand{\N}{{\set{N}}}
\newcommand{\bN}{\xoverline{\set{N}}}

\newcommand{\C}{{\set{C}}}

\newcommand{\Ps}{{\set{P}}}
\newcommand{\Qs}{{\set{Q}}}

\newcommand{\Ss}{{\set{S}}}

\newcommand{\A}{{\set{A}}}

\newcommand{\operator}[1]{\mathrm{#1}}

\newcommand{\RSRP}{\operator{RSRP}}
\newcommand{\HOL}{(\operator{HOL})}
\newcommand{\HOE}{(\operator{HOE})}
\newcommand{\HOW}{(\operator{HOW})}
\newcommand{\HOPP}{(\operator{HOPP})}
\newcommand{\HO}{(\operator{HO})}
\newcommand{\ca}{\operator{(C)}}
\newcommand{\cpa}{\operator{(CP)}}
\newcommand{\Asc}{\A^{\ca}}
\newcommand{\Ascp}{\A^{\cpa}}
\newcommand{\Ssc}{\Ss^{\ca}}
\newcommand{\Sscp}{\Ss^{\cpa}}
\newcommand{\Lp}{L^{(\operator{p})}}
\newcommand{\Lq}{L^{(\operator{q})}}
\newcommand{\Lr}{L^{(\operator{\rho})}}
\newcommand{\Lps}{L^{(\operator{\psi})}}
\newcommand{\symlog}{\operator{symlog}}
\newcommand{\symexp}{\operator{symexp}}
\newcommand{\sign}{\operator{sign}}

\makeatletter
\newsavebox\myboxA
\newsavebox\myboxB
\newlength\mylenA
\newcommand*\xoverline[2][0.75]{%
    \sbox{\myboxA}{$\m@th#2$}%
    \setbox\myboxB\null% Phantom box
    \ht\myboxB=\ht\myboxA%
    \dp\myboxB=\dp\myboxA%
    \wd\myboxB=#1\wd\myboxA% Scale phantom
    \sbox\myboxB{$\m@th\overline{\copy\myboxB}$}%  Overlined phantom
    \setlength\mylenA{\the\wd\myboxA}%   calc width diff
    \addtolength\mylenA{-\the\wd\myboxB}%
    \ifdim\wd\myboxB<\wd\myboxA%
       \rlap{\hskip 0.5\mylenA\usebox\myboxB}{\usebox\myboxA}%
    \else
        \hskip -0.5\mylenA\rlap{\usebox\myboxA}{\hskip 0.5\mylenA\usebox\myboxB}%
    \fi}
\makeatother

\begin{document}    
\title{Compositional Learning for Modular Multi-Agent Self-Organizing Networks}
\author{
\IEEEauthorblockN{Qi Liao\IEEEauthorrefmark{1}, 
			   Parijat Bhattacharjee\IEEEauthorrefmark{2}}
\IEEEauthorblockA{ 
	\IEEEauthorrefmark{1}Nokia Bell Labs, Stuttgart, Germany\\
	\IEEEauthorrefmark{2}Nokia, Bengaluru, India\\
E-Mail:
\IEEEauthorrefmark{1}\url{qi.liao@nokia-bell-labs.com}, 
	   \IEEEauthorrefmark{2}\url{parijat.bhattacharjee@nokia.com}} \vspace{-0.4in}
}

\maketitle
%\tableofcontents

\begin{abstract}
Self-organizing networks face challenges from complex parameter interdependencies and conflicting objectives. This study introduces two compositional learning approaches—Compositional Deep Reinforcement Learning (CDRL) and Compositional Predictive Decision-Making (CPDM)—and evaluates their performance under training time and safety constraints in multi-agent systems. We propose a modular, two-tier framework with cell-level and cell-pair-level agents to manage heterogeneous agent granularities while reducing model complexity. Numerical simulations reveal a $37.2\%$ reduction in handover failures, along with improved throughput and latency, outperforming conventional multi-agent deep reinforcement learning approaches. The approach also demonstrates superior scalability, faster convergence, higher sample efficiency, and safer training in large-scale self-organizing networks.
\end{abstract}

\makeatletter{\renewcommand*{\@makefnmark}{}\footnotetext{This work was supported by the German Federal Ministry of Education and Research (BMBF) project 6G-ANNA grant 16KIS077K.}\makeatother}

% \begin{IEEEkeywords}
% Compositional learning, modular deep learning, multi-agent deep reinforcement learning, self-organizing networks.
% \end{IEEEkeywords}

\section{Introduction}\label{sec:Introduction}
% AI/ML's application in SON
%\ac{SON} is one of the key enablers to autonomous networks, by leveraging the autonomy mechanisms for various functions such as \ac{MRO} and \ac{MLB} \cite{TS32500}. These \ac{SON} functions dynamically optimize network-level control parameters based on a wealth of \acp{KPI}. However, ensuring their practicality requires addressing several challenges. The strong interactions between diverse \ac{SON} functions often lead to potential disruptions, demanding careful inter-function synergy. Additionally, \ac{SON} mechanisms must be highly scalable to support deployment in large-scale networks with tens of thousands of cells. Finally, robust online safety assurance mechanisms are essential to maintain operations within predefined bounds.
\ac{SON} is a key enabler of autonomous networks, leveraging mechanisms like \ac{MRO} and \ac{MLB} \cite{TS32500} to dynamically optimize network control parameters using \acp{KPI}. However, its practicality hinges on overcoming challenges such as managing disruptions from inter-function interactions, ensuring scalability for large-scale networks, and implementing robust online safety mechanisms to maintain operational bounds.
% AI/ML for NG-RAN

% challenge of scalability and safe learning --> modular deep learning and transfer learning approaches

%\subsection{Related Works}
% state of the art: 
% - AI/ML especially deep reinforcement learning 
Deep learning-based methods have become instrumental in capturing the complexities of multi-objective functions in \ac{SON} \cite{Banerjee23}. For example, in \cite{mohajer2020mobility}, a deep learning-based mobility-aware \ac{MLB} solution was developed, tailoring \ac{HO} parameters to specific mobility patterns within individual cells. As networks evolve dynamically, many studies have turned to reinforcement learning for on-the-fly optimization. In \cite{mari2021service}, the authors aimed at improving cell edge \ac{QoE} while optimizing successful handover rates using Q-Learning. To further enhance model accuracy without escalating complexity, \ac{MADRL} algorithms are increasingly favored across various use cases. In \cite{Hu2022InterCellReDRL}, a coordinated multi-agent deep reinforcement learning algorithm was developed for capacity and \ac{QoE} optimization, in the context of slicing resource management. 

However, the efficacy of conventional \ac{MADRL} approaches heavily relies on domain-specific characteristics among agents, posing challenges in reproducibility and modularity when deploying solutions across diverse network scenarios and tasks, and often suffering from slow convergence and instability in large-scale systems. Emerging \ac{AI} techniques such as compositional learning \cite{purushwalkam2019task} and modular deep learning \cite{pfeiffer2023modular} are being introduced to address issues of poor model reproducibility and limited sample efficiency. 

In this study, we aim to tackle scalability, sample efficiency, and training safety issues in the context of multi-objective \ac{SON} functions in multi-agent network systems. Our contributions are listed in below.

\emph{1) Two-tier design for heterogeneous agents}: Existing \ac{MADRL}-based approaches typically define an agent for each cell, focusing on optimizing cell-specific parameters \cite{feriani2021single}. However, in mobility-related use cases such as \ac{MRO} and \ac{MLB}, parameters can be either cell-specific or cell-pair-specific (e.g., specific to cell edges) \cite{liao2022knowledge}. These parameters interact strongly, influencing neighboring cells' performance. To address this complexity, we decompose the global problem into two tiers: cell-level agents and cell-pair-level agents, capturing inter-agent dependencies through collaborative observations. To the best of our knowledge, this paper is the first to introduce cell-pair-level agents, with rewards carefully aligned with those of cell-level agents. This design enables a more effective representation of the interdependencies arising from \ac{HO} events between cell pairs, significantly improving coordination and performance.

\emph{2) Compositional learning and decision making}: Multiple \ac{SON} objectives are jointly influenced by a common set of features. We first developed the \ac{CDRL} method, which decomposes the reward function into sub-modules, enhancing the model's modularity and reusability while significantly accelerating convergence. However, in real-world systems, \ac{CDRL} often faces challenges of risky exploration and instability, making it more complex than supervised learning. To address this, we further propose \ac{CPDM}, which optimizes the same objectives as \ac{CDRL} but is simpler to implement and easier to train. Furthermore, we employed advanced deep learning techniques such as centralized training decentralized execution, order-agnostic sample augmentation, and symlog normalization to improve training performance. 

 \emph{3) Case study}: We evaluated the performance of the proposed approaches in optimizing multiple \ac{SON} objectives related to network mobility, capacity, and reliability using a system-level simulator. Compared to the conventional \ac{MADRL}, our approach exhibited a substantial $37.2\%$ reduction in total handover failures, coupled with enhancements in throughput and reduction in radio link failures, with ensured training safety within the constrained training time. Furthermore, we demonstrate that the proposed \ac{CPDM} method offers more effective training than \ac{CDRL}. 

The rest of the paper is organized as follows. In Section \ref{sec:SysModel}, we define the system model and problem formulation.  The \ac{CDRL} and \ac{CPDM} solutions are described in Section \ref{sec:PBDM}. In Section \ref{sec:Results}, we present the case study and the numerical results, and the paper is concluded in Section \ref{sec:Conclusion}.

\section{System Model and Problem Formulation}\label{sec:SysModel}
We consider a \ac{SON} system, comprising a set of $N$ network entities, such as cells, denoted by $\N$. Each cell $n\in\N$ has a set of $\bN_n$ neighboring cells, with cardinality $\left|\bN_n\right|=N_n$. Let the set of cell pairs associated with cell $n$ be denoted by $\C_n:=\{(n,m):m\in\bN_n\}$, where $(n,m)$ represents the cell edge between cells $n$ and $m$, and denote the whole set of cell pairs by $\C$. 
%The state of a cell is strongly dependent on the states and actions of its neighboring cells, due to the cell pair-wise \ac{HO} parameters and inter-cell interference. 
Note that a network entity is not confined to physical cells; it can also encompass virtual entities like network slices or network agents handling groups of users.
%, e.g., with the same mobility classes.  %Within the \ac{3GPP}, specific control parameters and \acp{KPI} are outlined for each \ac{SON} use case. 
For detailed information on the parameters and \acp{KPI} relevant to \ac{SON} use cases, we refer readers to \cite{TS32500}, and Section \ref{sec:Results} will explain those pertinent to our case study. 

{\bf Control parameters}. There are two types of control parameters: 1) {\bf Cell-level parameters}, such as antenna transmit power, are specific to individual cells. Assuming a number of $\Lp$ distinct cell-level parameters, denote them as $\vp_n:=[p_{n,1}, \ldots, p_{n,\Lp}]$ for cell $n\in\N$, where $p_{n,l}\in\Ps_l$ and $\Ps_l$ is the bounded space of the $l$-th cell-level parameter. Let $\vp:=[\vp_1, \ldots, \vp_N]\in \Ps$ be the global cell-specific parameters. 2) {\bf Cell pair-level} parameters,  often associated with \ac{HO} behaviors between two neighboring cells, including parameters like \acp{CIO}\cite{TS32500}. For a cell pair $(n,m)\in\C$, denote the collection of $\Lq$ parameters as $\vq_{n, m} := [q_{n,m,1}, \ldots, q_{n, m, \Lq}]$, where $q_{n,m,l}\in\Qs_l$. Here, $\Qs_l$ is the bounded space of the $l$-th cell pair-level parameter. Define $\vq:=[\vq_{n,m}: n\in\N, m\in\xoverline{\N}_n]\in \Qs$ as the global cell pair-level parameters. Note that the cell pair-level parameters are {\bf directional}, meaning $q_{n,m}$ may not equal $q_{m,n}$.

{\bf Observable \acp{KPI}}. The \ac{OAM} periodically gathers \acp{KPI} from all cells, typically averaged over intervals like every $5$ to $15$ minutes. Similar to control parameters, \acp{KPI} can be collected per cell, like cell traffic load or throughput, or per cell pair, including the \ac{HO} metrics between neighboring cells. Let the {\bf cell-level \acp{KPI}} for cell $n\in\N$ be denoted by $\vrho_n:=[\rho_{n,1}, \ldots, \rho_{n,\Lr}]\in\R^{\Lr}$, and the global cell-level \acp{KPI} be $\vrho:=[\vrho_1, \ldots, \vrho_N]$. Denote the  {\bf cell pair-level \acp{KPI}} for cell pair $(n,m)$ as $\vpsi_{n,m}:=[\psi_{n,m,1}, \ldots, \psi_{n,m,\Lps}]\in\R^{\Lps}$, and the global cell pair-level \acp{KPI} as $\vpsi:=\left[\vpsi_{n,m}: (n,m)\in\C\right]\in\R^{\Lps\sum_{n\in\N}N_n}$.

{\bf Markov decision process}. The system runs on discrete time intervals $t\in\NN_0$. Considering the network dynamics, we model the multi-cell system as a \ac{MDP} defined by tuple $\left\{\Ss, \A, P(\cdot), r(\cdot), \gamma\right\}$, where $\Ss$ denotes the state space, $\A$ is the action space, $P:\Ss\times\A\times\Ss\to [0,1]$ indicates the transition dynamics by a conditional distribution that an action in a state leading to a next state, the reward function is represented by $r:\Ss\times\A\to\R$, and $\gamma\in[0,1]$ is the discount factor.  

At time $t$, we denote the {\bf global state} as $\vs(t):=[\vrho(t), \vpsi(t)]\in\Ss:=\R^{N\Lr\Lps \sum_{n\in\N}N_n}$, as an observation of the entire system consisting of both cell-level and cell pair-level \acp{KPI}. % At each cell $n\in\N$, the local observation is denoted by $\vs_n(t):=[\vrho_n(t), \vpsi_{n,1}, \ldots, \vpsi_{n,N}]\in\Ss_n:=\R^{\Lr\Lps(N-1)}$, including both the per cell and per cell pair \acp{KPI}.
The {\bf global action} is denoted by $\va(t):=[\vp(t), \vq(t)]\in\A:=\Ps\times\Qs$. The {\bf global reward} $r(t)\in\R$, dependent on $(\vs(t),\va(t))$, is a scalar computed from the collected \acp{KPI} $\vrho(t+1)$ and $\vpsi(t+1)$. 
%For instance, typical multi-objective function can be a weighted sum of all interested \acp{KPI}.

In line with the classical reinforcement formulation, the global problem is finding the policy $\pi:\Ss\to\A$ for optimizing the control parameters based on the \ac{KPI} observations,  to maximize the expected cumulative discounted reward of a trajectory for a finite time horizon $T$:
\begin{equation}
\max_{\pi}\Ex_{\pi}\left[\sum_{t=0}^T\gamma^t r(\vs(t), \va(t))\right], s.t.~\va(t)\in\A.
\label{eqn:glob_prob}
\end{equation}
% \begin{problem}[Global RL problem]
% \label{prob:global}
% \begin{equation}
% \max_{\pi}\Ex_{\pi}\left[\sum_{t=0}^T\gamma^t r(\vs(t), \va(t))\right], s.t.~\va(t)\in\A.
% \label{eqn:glob_prob}
% \end{equation}
% \end{problem}

Given the vast state and action space, the global problem is intractable. Therefore, we seek distributed and efficient solutions with online safety assurance.

\section{Compositional Learning and Decision-Making}\label{sec:PBDM}
In this section, we propose a novel distributed compositional learning and decision-making approach to accelerate learning and guarantee training safety. The two key enablers are: 1) \emph{two-tier modular design for heterogeneous agents}, and 2) \emph{compositional learning of the value function}. 
\subsection{Two-Tier Modular Design for Heterogeneous Agents}\label{ssec:localagents}
 We are interested in decomposing the global agent into local ones for efficient and highly scalable distributed implementation. 
%  The key challenges include:
% \begin{itemize}
%     \item Balancing \emph{model complexity} and \emph{approximation accuracy}.
%     \item Improve \emph{reusability and transferability} of the learned models within local agents, facilitating their transfer and reuse among similar agents. 
% \end{itemize}
Many works have proposed constructing per-cell agents to optimize cell-level parameters $\vp_n$ based on local observations $\vrho_n$ and possibly exchanged observations  \cite{Hu2022InterCellReDRL}. 
%Some works have improved the approximation accuracy by expanding the observation space through information exchange with neighboring cells \cite{Hu2022InterCellReDRL}. 
However, when considering cell pair-level parameters and \acp{KPI},  the challenge arises from both strong inter-cell dependencies and  {\bf varying number of neighboring cells} each cell may have, leading to different state and action dimensions across cells. This requires local agents to adjust their model size according to the number of neighboring cells, with limited model reusability. 
%Also, if each local agent must adjust the model size based on the number of neighboring cells, it may only train with its own collected samples, leading to significantly low sample efficiency and prolonged convergence times.

\begin{figure}[t]
     \centering     \includegraphics[width=0.48\textwidth]{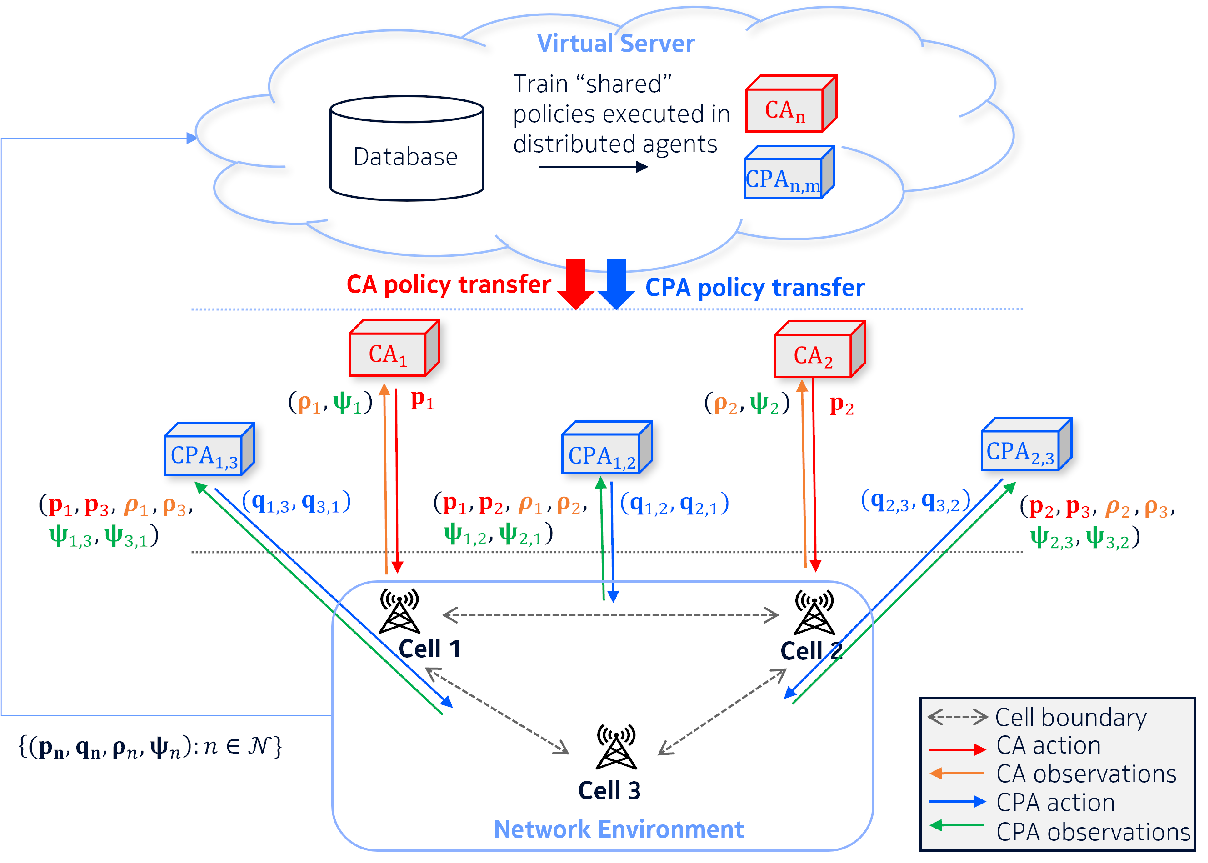}
         \caption{\ac{CTDE}-based two-tier agents.}
\label{fig:TraingAgents}
\end{figure}

We propose a two-tier approach, comprising the first-tier \acp{CA} and the second-tier \acp{CPA}, each with fixed input and output dimensions universally applicable across different cells and cell pairs, as shown in Figure \ref{fig:TraingAgents}. We adopt the widely used \emph{\ac{CTDE}} approach \cite{lowe2017multi}, allowing local samples from different cells to contribute to training a shared model that captures general behaviors across various cells. Then, during the inference phase, the trained model can be loaded and executed distributedly. Note that \acp{CA} and \acp{CPA} do not have to be deployed at their respective cells, but anywhere in a virtual server.
%, such as the non-real-time RAN Intelligent Controller (RIC) in Open-RAN \cite{polese2023understanding}. 
%The \ac{CTDE} approach significantly improves the sample efficiency and learning rates. 
%We define \acp{CA} and \acp{CPA} in below.
%
\subsubsection{Cell-Level Agents}\label{sssec:CA}
At time $t$, the {\bf action} of the \ac{CA} in the $n$-th cell is $\va_n(t):=\vp_n(t)$. The {\bf state} is defined as $\vs_n(t):=[\vrho_n(t), \vg_0(\vpsi_n(t))]$, where $\vg_0$ is a mapping from $\Lps N_n$ cell pair \acp{KPI} (varying dimensions for each $n\in\N$) to a reduced fixed dimension of features. For instance, an efficient and practical solution is to simply compute the mean of each cell pair \ac{KPI} over cell pairs $\C_n$ \cite{Hu2022InterCellReDRL}:
\begin{equation}
  \label{eqn:cellpair_info}
\begin{aligned}
  \vg_0  : \R^{\Lps N_n}  \to\R^{\Lps} &:\vpsi_n \mapsto \left[\overline{\psi}_{n,l}:l= 1, \ldots, \Lps\right] \\
  \mbox{where }  \overline{\psi}_{n,l} &:=\frac{1}{N_n}\sum\nolimits_{m\in\xoverline{\N}_n}\psi_{n,m,l}.
\end{aligned}
\end{equation}
Let the action and state spaces of any \ac{CA} be denoted by $\Asc$ and $\Ssc$, respectively. We have $\Asc:=\prod_{l=1}^{\Lp}\Ps_l$ and $\Ssc:=\R^{\Lr + \Lps}$, with fixed action and state dimensions for any $n\in\N$, regardless of $N_n$.

The multi-objective {\bf reward} captures both cell \acp{KPI} $\vrho_n$ and cell pair \acp{KPI} $\vpsi_n$. 
%Depending on the desired performance, operators can customize different metrics to map $\vrho_n$ and $\vpsi_n$ into scalars, respectively. 
Detail definitions will be provided in Section \ref{ssec:casestudy} for the case study. Here we give a general formulation:
\begin{equation}
\begin{aligned}
    & r_n(t)  =  w_1 g_1\big(\vrho_n(t+1)\big) +\\
    & w_2 \frac{1}{N_n} \sum\nolimits_{m\in\bN_n}g_2\big(\vpsi_{n, m}(t+1), \vpsi_{m,n}(t+1)\big),
\end{aligned}
\label{eqn:rwd_CA}
\end{equation}
where $g_1: \R^{\Lr}\to\R$ and $g_2: \R^{2\Lps}\to\R$, and $w_1$ and $w_2$ are the weight factors balancing cell \acp{KPI} and  cell pair \acp{KPI}. 
\subsubsection{Order-Agnostic Cell Pair-Level Agents}\label{sssec:CPA}
The action of the \ac{CPA} for cell pair $(n,m)$ is $\vba_{n,m}(t):=[\vq_{n,m}(t), \vq_{m,n}(t)]\in\Ascp:=\prod_{l=1}^{\Lq}\Qs_l^2$. We allow the \ac{CPA} to observe both its cell pair \acp{KPI} and its belonging cell's \acp{KPI}. In addition, we include the cell parameters $\vp_n$ in the observation space, as the first-tier \acp{CA} make the decision first. Thus, we have $\vbs_{n,m}(t):=[\vp_n(t), \vp_m(t), \vrho_n(t), \vrho_m(t), \vpsi_{n,m}(t), \vpsi_{m,n}(t)]$. The state space yields $\Sscp:=\prod_{l=1}^{\Lp}\Ps_l^2\times\R^{4\Lr\Lps}$, and the state has a fixed dimension of $2(\Lp + \Lr + \Lps)$. 
We further compute the reward with the same mapping $g_1$ and $g_2$: 
\begin{equation}
\begin{aligned}
    \br_{n,m}(t) = & w_1 g_1\big(\vrho_n(t+1)\big) + \\
    & w_2 g_2\big(\vpsi_{n, m}(t+1), \vpsi_{m,n}(t+1)\big).
    \end{aligned}
    \label{eqn:rwd_CPA}
\end{equation}
\begin{Rem}[Aligned reward of \acp{CA} and \acp{CPA}]
\label{rem:rwd_align}
From \eqref{eqn:rwd_CA} and \eqref{eqn:rwd_CPA}, we have $r_n(t)=(1/N_n)\cdot\sum_{m\in\xoverline{N}_n} \br_{n,m}(t)$. Thus, improving the rewards of \acp{CPA} associated to cell $n$ also improves the reward of the \ac{CA} of cell $n$. 
\end{Rem}
\begin{table*}[ht]
\begin{center}
\caption{Comparison among the agents and their dimensions.}
\label{tab:agent_compare}
\begin{tabular}{|l|l|ll|}
\hline
\multirow{2}{*}{} & \multirow{2}{*}{\begin{tabular}[c]{@{}l@{}}Conventional neighbor-aware \\ per-cell agent\end{tabular}} & \multicolumn{2}{l|}{Our proposed two-tier agents}               \\ \cline{3-4} 
                  &                                                                                           & \multicolumn{1}{l|}{CA}              & CPA                      \\ \hline
Action            & $\va'_n :=[\vp_n, \vq_n]$                                                    & \multicolumn{1}{l|}{$\va_n:=\vp_n$}      & $\vba_{n,m}=[\vq_{n,m}, \vq_{m,n}]$ \\ \hline
Action dim.       & $\Lp + N_n \cdot \Lq$                                                                           & \multicolumn{1}{l|}{$\Lp$}           & $2\Lq$                 \\ \hline
State             & $\vs_n':=[\vrho_n, (\vpsi_{n,m}, \vpsi_{m,n}): m\in\bN_n]$                                            & \multicolumn{1}{l|}{$\vs_n:=[\vrho_n, \ve{g}_0(\vpsi_n)]$}           & $\vbs_{n,m}:=[\vp_n, \vp_m, \vrho_n, \vrho_m, \vpsi_{n,m}, \vpsi_{m,n}]$                \\ \hline
State dim.        & $\Lr + 2 N_n \Lps$                                                                            & \multicolumn{1}{l|}{$\Lr + \Lps$}    & $2(\Lp + \Lr + \Lps)$            \\ \hline
Reward            & $r_n\in\R$ in \eqref{eqn:rwd_CA}                                                                              & \multicolumn{1}{l|}{$r_n\in\R$ in \eqref{eqn:rwd_CA}} & $\br_{n,m}\in\R$ in 
 \eqref{eqn:rwd_CPA} \\ \hline
\end{tabular}
\end{center}
\vspace{-2ex}
\end{table*}
\begin{Rem}[Order-agnostic training]
\label{rem:order_agnostic}
A \ac{CPA} should represent the joint behavior of both cell boundaries ($n\to m$ and $m\to n$) within an order-agnostic cell tuple $(n,m)$. For example, for cell pair $(1,2)$, it should not matter whether $n=1, m=2$ or $n=2, m=1$. Therefore, 
to ensure \acp{CPA} adaptable to arbitrary orders, we augment collected sample sets, including in the replay buffer both samples: $\big(\vbs_{n,m}(t), \vba_{n,m}(t), \br_{n,m}(t), \vbs_{n,m}(t+1)\big)$ and $\big(\vbs_{m,n}(t), \vba_{m,n}(t), \br_{m,n}(t), \vbs_{m,n}(t+1)\big)$. 
\end{Rem}

Compared to the conventional per-cell agents that integrate all cell pair parameters and \acp{KPI} into action and state spaces, our two-tier approach achieves much lower complexity and greater modularity, while aligning the objectives of \acp{CA} and \acp{CPA}. 
%Additionally, fixed dimensions of the two-tier models enhance modularity, allowing learned models to be transferred to other systems.
Table \ref{tab:agent_compare} compares the dimensions of our proposed two-tier agents with conventional per-cell agents that include $\vq_n$ in action and $\{(\vpsi_{n,m}, \vpsi_{m,n}):m\in\xoverline{\N}_n\}$ in states.

\subsection{Actor-Critic Algorithm and Alternating Direction Method}
\label{ssec:ActorCritic}
With the defined states, actions, and rewards, the two-tier \acp{CA} and \acp{CPA} search for the policies $\pi_n:\Ssc\to\Asc$, $\forall n\in\N$ and $\xoverline{\pi}_{n,m}:\Sscp\to\Ascp$, $\forall (n,m)\in\C$, respectively, by jointly solving the following \ac{CA} and \ac{CPA} problems:
\begin{align}
\max_{\pi_n:\va_n\in\Asc} & \Ex_{\pi_n}\left[\sum_{t=0}^T\gamma^t r_n(\vs_n(t), \va_n(t))\right],  \label{eqn:CA_problem}\\
\max_{\substack{\xoverline{\pi}_{n,m}: \\ \vba_{n,m}\in\Ascp}} &\Ex_{\xoverline{\pi}_{n,m}}\left[\sum_{t=0}^T\gamma^t \br_{n,m}(\vbs_{n,m}(t), \vba_{n,m}(t))\right].
\label{eqn:CPA_problem}
\end{align}

\subsubsection{Actor-Critic Algorithms}
In subsequent subsections, for brevity, we refer to the state and action at time $t$ as $\vs_t, \va_t$ respectively, while $r, \pi$ represent the reward and policy, respectively, regardless of \ac{CA} or \ac{CPA}. 
%We may also use $\vs_t, \va_t$ and $\vs(t), \va(t)$ interchangeably. 
We choose actor-critic methods \cite{Konda1999ActorCriticA} to solve \ac{DRL} problems, for its effectiveness when dealing with high dimensional spaces. Such methods learn a critic function $Q(\vs_t, \va_t|\vthe)$, typically a \ac{MLP} parameterized by $\vthe$, to approximate the value function $Q^{\pi}(\vs_t, \va_t)$, which describes the expected return after taking an action $\va_t$ in state $\vs_t$ following policy $\pi$. In addition, an actor $\pi(\vs_t|\vphi)$ parameterized by $\vphi$ updates the action $\va_t$ based on observation $\vs_t$. 
In this paper, we consider the widely used \ac{TD3} \cite{Fujimoto2018AddressingFA}. 
The action is updated by policy gradient based on the expected cumulative reward $J$ with respect to the actor parameter $\vphi$, as:
\begin{equation}
	\nabla_{\vphi}J \approx \Ex\left[\nabla_{\va}Q(\vs, \va|\vthe)|_{\vs=\vs_t, \va=\pi(\vs_t)}\nabla_{\vphi}\pi(\vs_t|\vphi)\right].
 \end{equation}
 The critic parameter $\vthe$ is updated by minimizing the loss of \ac{TD}, given by:
 \begin{equation}
 	L_t\left(\vthe\right) = \Ex\left[\left(g_t - Q(\vs_t, \va_t|\vthe)\right)^2\right], \\
  \label{eqn:TD_loss}
 \end{equation}
where $g_t = r_t + \gamma Q\left(\vs_{t+1}, \pi\left(\vs_{t+1}|\vphi\right)|\vthe\right)$.
\subsubsection{Joint \ac{CA} and \ac{CPA} Training with Alternating Direction} The classical actor-critic methods can solve the local problems in \acp{CA} and \acp{CPA} if they operate independently. However, the rewards in \ac{CA} and \ac{CPA} are jointly influenced by cell-level and cell pair-level actions.
To address this interaction, one potential approach is to employ alternating direction methods, which iteratively update different sets of parameters, while holding others fixed. In the context of training \acp{CA} and \ac{CPA}, this involves alternating between updating the actors and critics in \acp{CA} and the actors and critics in \acp{CPA}, as well as alternating the actions $\va_n$ and $\vba_{n,m}$. Because the objectives of \acp{CA} and \acp{CPA} are aligned (Remark \ref{rem:rwd_align}), doing this allows for addressing the interplay between these two levels of automata. Additionally, given the broader impact of \ac{CA} decisions on associated cell pairs, we use a lower update frequency for \acp{CA} compared to \acp{CPA} to promote stability during training and reduce the risk of over-fitting. Note that although \ac{ADMM} exhibits good performance for convex problems \cite{boyd2004convex}, its convergence in deep learning remains intriguing. Empirical evidence suggests that they can converge effectively in deep learning \cite{zeng2021admm}.
%the theoretical understanding of this behavior is not as well-established.
%, due to the nonlinear, nonconvex nature of deep learning problems. 
In this paper, we only provide empirical studies to show its effectiveness in joint training of \acp{CA} and \acp{CPA}.
\subsection{CDRL: Compositional Deep Reinforcement Learning}\label{ssec:CDRL}
\begin{figure}[t]
     \centering     \includegraphics[width=0.48\textwidth]{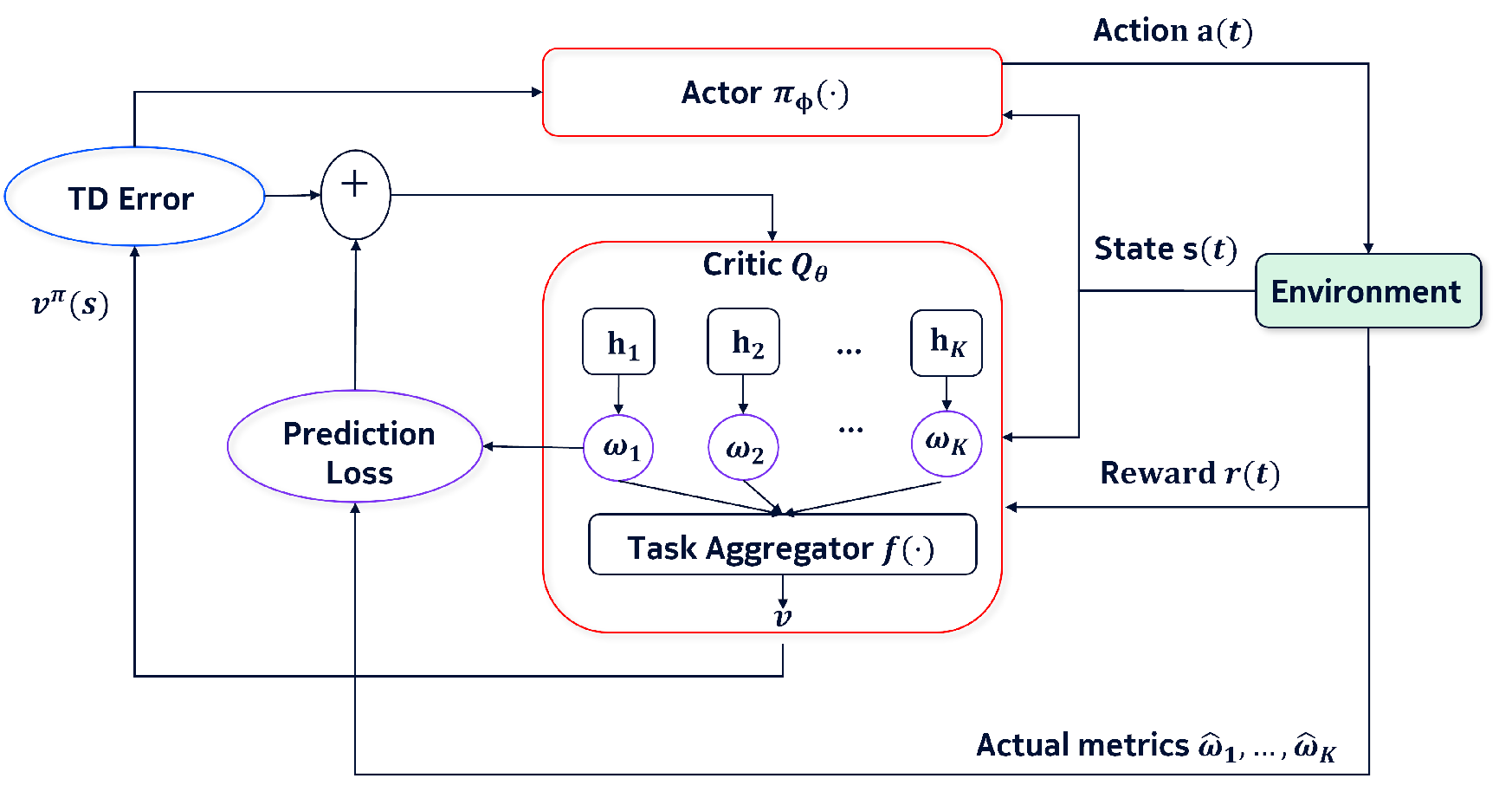}
         \caption{Basic architecture of \ac{CDRL}.}
\label{fig:CDRL}
\end{figure}

The above-introduced actor-critic algorithms often face challenges of slow convergence, especially with complex, multi-objective reward functions. Considering that \ac{SON} multi-objectives are composed of different aspects of the performance, e.g., enhancing throughput, reducing delay and \ac{HO} failures, we propose to decompose the critic into multiple sub-critics, each responsible for evaluating a different objective. By breaking down the value function into smaller, more specialized components, we can enhance the modularity, interpretability, and hierarchical representation of the \ac{DRL} model. As shown in Figure \ref{fig:CDRL}, the value function to estimate $v^{\pi}(s)$ is decomposed to $K$ prediction functions of the sub-critics $\{\vh_1, \ldots ,\vh_K\}$, with parameters $\{\vthe_1, \ldots, \vthe_K\}$, providing predicted metrics $\{\ve{\omega}_1, \ldots, \ve{\omega}_K\}$ respectively. Then, a task aggregator $f$, e.g., an \ac{MLP} with parameters $\vthe_0$,  returns the Q value based on the predicted metrics. The loss of the critic includes both the \ac{TD} error and the prediction loss:

\begin{equation}
\begin{aligned}
\tilde{L}_t(\vthe_0, \vthe_1, & \ldots, \vthe_K) :=  L_t(\vthe_0,, \vthe_1, \ldots, \vthe_K) + \\
& \frac{1}{K}\sum_{k=1}^K \left\|\vh_k(\vs_t, \va_t|\vthe_k)-\hat{\ve{\omega}}_{k, t+1}\right\|_2,
\end{aligned}
\label{eqn:compositional_loss}
\end{equation}

where $L_t$ is the classical \ac{TD} loss at time $t$ of \ac{TD3} algorithm, $\hat{\ve{\omega}}_{k, t+1}$ is the actual measured $k$-th metrics at time $t+1$.
%, and $\left\|\cdot\right\|_2$ denotes L2 norm.  

%We outline the joint training of the general \ac{CA} and \ac{CPA} with \ac{CDRL} in Algorithm \ref{algo:CDRL}. 
% The \acp{CA} and \acp{CPA} are jointly trained with alternating direction. During the training phase, samples from various cells and cell pairs collectively contribute to training a unified \lq\lq general\rq\rq \ \ac{CA} and \ac{CPA}. In contrast, during the inference or execution phase, the trained \ac{CA} and \ac{CPA} can be loaded distributedly to infer the local actions $\vp_n$ and $[\vq_{n,m}, \vq_{m,n}]$, respectively.
% \begin{algorithm}
% \caption{Joint training of \ac{CA} and \ac{CPA} with \ac{CDRL}.}
% \label{algo:CDRL}
% \begin{algorithmic}[1]
% \State \textbf{Initialize:} \\
% \ac{CA}: actor $\pi$, update interval $\tau$, replay buffer $\B$\\
% \ac{CPA}:  actor $\bpi$, update interval $\btau$, replay buffer $\bB$
% \For{$t\in T$}
% \State Collect $\{(\vp_n(t), \vq_n(t), \vrho_n(t), \vpsi_n(t)):n\in\N\}$.   
% \EndFor
% \end{algorithmic}
% \end{algorithm}

\begin{Rem}[Symlog normlization] 
The scale of the network parameters and \acp{KPI} can vary significantly between domains. Predicting large targets using a squared loss function may result in divergence, while absolute and Huber losses might impede learning progress. To address the dilemma, we use \emph{symlog} normalization and apply a function selected from the bi-symmetric logarithmic family: 
$\symlog(x):=\sign(x)\ln(|x| + 1)$. 
To read out the prediction and compute the loss function of the prediction, the inverse transformation is $\symexp(x) := \sign(x)(\exp(|x|)-1)$.
% \begin{equation}
% \symlog(x):=\sign(x)\ln(|x| + 1).
% \label{eqn:symlog}
% \end{equation}
% To read out the prediction and compute the loss function of the prediction, the inverse transformation is 
% \begin{equation}
% \symexp:= \sign(x)(\exp(|x|)-1). 
% \end{equation}
\label{rem:symlog_norm}
\end{Rem}
\subsection{CPDM: Compositional Predictive Decision-Making}\label{ssec:CPDM}
Despite the benefits of the proposed two-tier mechanism and compositional learning, challenges persist in balancing exploration and exploitation in \ac{DRL} approaches. The exploration phase often results in hazardous outcomes during training. In real-life scenarios, predictive decision-making approaches may offer advantages of safety guarantees compared to \ac{DRL} algorithms \cite{hu2023fast}.
% . For instance, in our prior work \cite{hu2023fast}, we discover that a hybrid deep learning approach, integrating deep learning-based prediction models with classical constrained optimization methods, facilitates a faster and safer training process. 
%Therefore, in this paper, we conduct a comparison between \ac{CDRL} and \ac{CPDM}, to address both academic curiosity and industrial relevance.

The proposed \ac{CPDM} approach modifies the action space by optimizing the step size of increasing or decreasing control parameters. Specifically, for cell-level parameters we have $\va_n(t)=\vp_n(t+1) - \vp_n(t)\in \Delta\Asc$ for $n\in\N$, and for cell pair-level parameters $\vba_{n,m}(t)=\vq_{n,m}(t+1) - \vq_{n,m}(t)\in\Delta\Ascp$, $\forall (n,m)\in\C$.
In practical implementations, the change in action at each step is restricted to ensure the stability of the system. Consequently,  $\Delta\Asc$ and $\Delta\Ascp$ are significantly smaller than $\Asc$ and $\Ascp$. 
%For example, Consider the  \ac{HO} parameter \ac{CIO}, which typically ranges from $-24$ dB to $24$ dB \cite{TS38331}, while the change action space can be limited to $\{-2, -1, 0, 1, 2\}$ dB.  %This constrained action space ensures that adjustments made to the parameters are within a safe and manageable range, facilitating stable and reliable operation of the system.

With the reduced action space, we have the ability to exhaustively compute the predicted value function for all potential changes within the constrained action space, and select the one with the maximum predicted value:  
\begin{align}
\va_n^{\ast} & =\argmax_{\va_n\in\Delta\Asc}f^{\ca}(\vs_n, \va_n),
\label{eqn:predictiveAction_cell}\\
\vba_{n,m}^{\ast} & =\argmax_{\vba_{n,m}\in\Delta\Ascp}f^{\cpa}(\vbs_{n,m}, \vba_{n,m}),
\label{eqn:predictiveAction_cellpair}
\end{align}
where $f^{\ca}$ and $f^{\cpa}$ are the aggregator functions of \ac{CA} and \ac{CPA} respectively.
This exhaustive search approach is facilitated by modern deep learning frameworks, which efficiently inference large batches of input samples.

\section{Case Study and Performance Evaluation}\label{sec:Results}
We apply the proposed \ac{CDRL} and \ac{CPDM} approaches to \ac{SON} use cases, targeting multi-objective optimization to enhance \ac{HO} performance, throughput, and latency while reducing \acp{RLF} by tuning mobility parameters such as \ac{TTT} and \ac{CIO}.
%We also evaluate and compare their effectiveness using a large-scale system-level simulator. 
%

\subsection{Case Study}\label{ssec:casestudy}
With a slight abuse of notation, for \ac{CA} $n$, we define a cell-level parameter \ac{TTT} $p_n$ within the space $\Ps:= \{40, 64, 80, 100, 128, 160, 256, 320, 480,  512, 640, 1024, 1280,\\ 2560, 5120\}$ in ms. For \ac{CPA} $(n,m)$, the cell pair-level parameters \acp{CIO} are $q_{n,m}, q_{m,n}\in\Qs:= \{-24, -23, \ldots, 24\}$ in dB \cite{TS38331}. Namely, we deal with the case where $\Lp=\Lq=1$.  

Although \ac{TTT} and \ac{CIO} are mobility-related parameters, they jointly affect various \acp{KPI} including \ac{HO} \acp{KPI}, cell load, throughput, and latency. This is because a user served by cell $n$ will undergo handover to the neighboring cell $m\in\bN_n$ if the following criterion holds  \cite{TS38331}:
\begin{equation}
\RSRP_m>\RSRP_n + q_{n,m} \mbox{ holds for } p_n \mbox{ ms},
\end{equation} 
where $\RSRP_i$ is the \ac{RSRP} from cell $i$. Conversely, similar conditions apply for handover from cell $m$ to cell $n$. 
Thus, large values of them may lead to too-late handover ({\bf HOL}), potentially resulting in \ac{RLF} before the handover completes. In contrast, small values may cause too-early handover ({\bf HOE}), where a handover occurs before the neighboring cell can provide a sustainable signal quality, leading to a
\ac{RLF} right after the handover. Another event due to the early \ac{HO} decisions is the wrong-cell handover ({\bf HOW}). Also, inappropriate settings of $q_{n,m}$ and $q_{m,n}$ lead to frequent handovers between cells, commonly known as ping-pong handover ({\bf HOPP}). Moreover, the handover behavior directly influences load distribution across cells, thereby substantially affecting cell throughput and latency. 

{\bf Two-tier actions.} When applying \ac{CDRL}, because \ac{TD3} works with continuous action space, the actions of \ac{CA} $n$  and \ac{CPA} $(n,m)$ are defined as $a_n:=p_n\in\Asc:=[40, 5120]$ and $\bar{a}_{n,m}:=[q_{n,m}, q_{m,n}]\in\Ascp:=[-24, 24]^2$, respectively. Then, we choose the nearest values to the actors' outputs in the defined sets $\Ps$ and $\Qs$ to execute. On the other hand, for \ac{CPDM}, we have different search space $\Delta\Asc(p_n(t))$, depending on the current \ac{TTT} value $p_n(t)$, because the values in $\Ps$ are not uniformly distributed. It computes the difference from the five values, spanning from the two on the left to the two on the right of $p_n(t)$ within the set $\Ps$, and $\Delta\Ascp:=\{-2, -1, 0, 1, 2\}^2$ in dB. Thus, we can search the whole discrete action space, and the size of the search space for each \ac{CA} is only $5$, and for each \ac{CPA} is $25$. The cell pair actions $\bar{a}_{n,m}$ are updated every $15$ minutes, while the update frequency of $a_n$ is slower, for every hour, allowing for more stable adjustments to cell-level parameters.

{\bf Partial-observable states.} The states for both \ac{CDRL} and \ac{CPDM} are defined in the same way as presented in Table \ref{tab:agent_compare}. The following $7$ cell-level \acp{KPI} are included in $\vrho_n$ for each \ac{CA} $n$: downlink average user throughput, uplink average user throughput, downlink \ac{PRB} usage, uplink \ac{PRB} usage, number of active users, number of \acp{RLF}, and average \ac{CQI}. The cell pair-level \acp{KPI} $\vpsi_{n,m}$ include the number of \ac{HO} attempts, \ac{HO} success ratio, and the numbers of HOLs, HOEs, HOWs, and HOPPs.  

{\bf Multi-objective rewards.} The multi-objective reward is constructed based on the following three prediction functions: \\
1) The {\bf \ac{HO} prediction} function $h_1(\cdot)$ is designed to predict \ac{HO} cost, defined as 
$C^{\HO} = \alpha_0 R^{\HOL} - \alpha_1 R^{\HOE} - \alpha_2 R^{\HOW} - \alpha_3 R^{\HOPP}$,
where $\alpha_i, i=0,\ldots, 3$ are positive weights in the range $[0,1]$, and $R^{\HOL}$, $R^{\HOE}$, $R^{\HOW}$, $R^{\HOPP}$ represent the ratios of \ac{HO} events (HOL, HOE, HOW, and HOPP) to the total number of HO attempts, respectively, typically set to $1, 1, 1, 0.2$. Since these ratios cannot exceed $1$, the \ac{HO} cost is in the range $[0,1]$. A large positive \ac{HO} cost indicates too-late \ac{HO} decisions, while a large negative value indicates too-early \ac{HO} decisions. Ideally, $C^{\HO}$ should approach $0$, indicating a balanced \ac{HO} decision strategy with a good trade-off between too-early and too-late \ac{HO} behaviors. \\
2) The {\bf throughput and latency classification} function $h_2(\cdot)$ is designed to predict the quantified class of throughput and latency, with three predefined classes \lq\lq good\rq\rq, \lq\lq normal\rq\rq, \lq\lq poor\rq\rq. This classification approach is chosen over predicting exact values because throughput and latency are influenced by dynamic user traffic patterns, and a low throughput during peak hours may not necessarily indicate poor performance. The labels for these classes are computed based on the time-dependent distribution of throughput and latency observations. This involves analyzing historical data to determine the typical range of throughput and latency values observed at different times of the day and define the thresholds to classify them into the appropriate class.\\
3) The {\bf \ac{RLF} anomaly prediction} function $h_3(\cdot)$ is a binary classifier to predict the anomaly increase of \ac{RLF} events. This choice of a binary classifier is motivated by the need to exclude \ac{RLF} caused by other network issues, such as poor coverage. Similar to the throughput and latency classifier, the labels are computed based on historical data analysis. 

To incorporate the \ac{HO} cost, throughput and latency classification, and \ac{RLF} anomaly prediction into rewards \eqref{eqn:rwd_CA} and \eqref{eqn:rwd_CPA}, we define auxiliary functions $g_1(\cdot)$ and $g_2(\cdot)$ as follows. $g_1(\cdot)$ computes $1-\left|C^{\HO}\right|$, ensuring it lies in the range $[0,1]$. $g_2(\cdot)$ computes $\1_{\{\hat{h}_2='good'\}} + 0.5 \cdot \1_{\{\hat{h}_2='normal'\}} + \1_{\{\hat{h}_3='normal'\}}$ where $\1_{\{X\}}$ is the indicator function equals $1$ if event $X$ occurs. Thus, the \ac{CA} and \ac{CPA} rewards are both upper-bounded by $3$. 
%This ensures that higher rewards are given when both throughput and latency are classified as 'good', and the  \ac{RLF} prediction indicates normal behavior.

%
\subsection{Numerical Results}\label{ssec:results}
\begin{table*}[ht!]
\begin{center}
\caption{\ac{KPI} comparison with $14$-day trial.}
\label{tab:KPI_compare}
\begin{tabular}{|l|l|l|l|l|l|l|l|l|}
\hline
      & \begin{tabular}[c]{@{}l@{}}DL throughput\\ in Mbps\end{tabular} & \begin{tabular}[c]{@{}l@{}}DL latency\\ in ms\end{tabular} & \begin{tabular}[c]{@{}l@{}}HOLR\\ in $\%$\end{tabular} & \begin{tabular}[c]{@{}l@{}}HOER\\ in $\%$\end{tabular} & \begin{tabular}[c]{@{}l@{}}HOWR\\ in $\%$\end{tabular} & \begin{tabular}[c]{@{}l@{}}HOPPR\\ in $\%$\end{tabular} & \begin{tabular}[c]{@{}l@{}}Throughput anomaly  \\ in $\%$\end{tabular} & \begin{tabular}[c]{@{}l@{}}RLF anomaly\\ in $\%$\end{tabular} \\ \hline
DFLT  & $29.62$ & $6.13$ & $0.93$ & $0.43$   & $0.39$   & $21.23$  & N/A  & N/A  \\ \hline
MADRL & $28.61$  &  $\mathbf{5.81}$  & $1.22$  & $0.36$  & $0.46$   &  $\mathbf{16.37}$   & $28.84$   & $0,63$  \\ \hline
H-MADRL & $29.23$  &  $6.02$  & $0.88$  & $0.38$  & $0.41$   &  $17.56$   & $16.95$   & $0,35$  \\ \hline
CDRL  & $30.05$                                                         & $5.95$                                                     & $0.72$                                               & $0.31$                                               & $0.38$                                               & $18.10$                                                 & $12.48$                                                             & $0,28$                                                            \\ \hline
CPDM  &  $\mathbf{30.14}$                                  & $5.90$                                                     & $\mathbf{0.62}$                        &  $\mathbf{0.30}$                        & $\mathbf{0.36}$                                               & $17.37$                                               & $\mathbf{6.26}$                                       &  $\mathbf{0.02}$                                     \\ \hline
\end{tabular}
\end{center}
\end{table*}
\begin{figure}[t]
      \centering     
\includegraphics[width=0.48\textwidth]{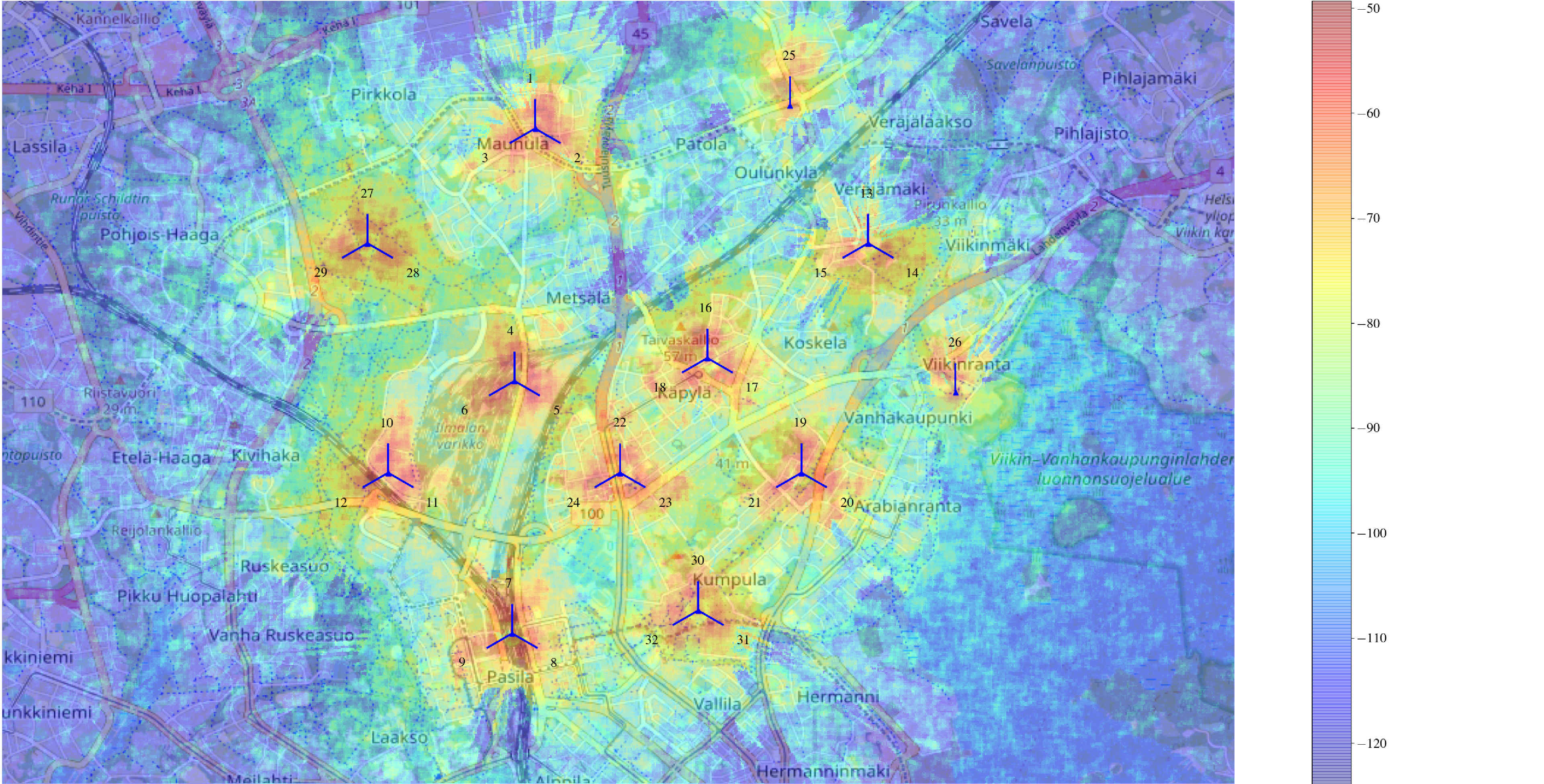}
    \caption{Simulated network scenario with 10 three-sector sites and 2 one-sector sites in Helsinki.}
 \label{fig:scenario}
 \end{figure}
We evaluate the proposed \ac{CDRL} and \ac{CPDM} approaches using a self-developed system-level simulator mimicking a real-world scenario. The radio map is extracted from a part of Helsinki, including $12$ sites with $32$ cells operating in the $2.4$ GHz frequency band as depicted in Figure \ref{fig:scenario}. 
The user's mobility is generated with the SUMO traffic simulator \cite{SUMO2018}. The number of users in the simulated area follows a realistic daily pattern based on real data. The agent interacts with the simulator with a granularity of simulation slots $900$, reflecting 15-minute intervals of real-life time. This allows for $96$ interactions per day. A total of $12$ days of training and $2$ days of evaluation are conducted, that is, with a maximum hour index of 336.

\subsubsection{Training Parameters}
As for the neural network architecture, for both \acp{CA} and \acp{CPA}, \acp{MLP} are used for the actor network, with $3$ hidden layers. The number of neurons for each layer is set to $(64, 32, 8)$. Additionally, each compositional critic predictor is implemented as an MLP with 3 hidden layers, having $(32, 16, 8)$ neurons. The symlog normalization described in Remark \ref{rem:symlog_norm} is applied to both inputs and outputs. The learning rates of
the actor and critic are $0.001$ and $0.002$, respectively. The
batch size is $64$, and the Adam optimizer is employed. 

% \begin{figure}[t]
%      \centering     \includegraphics[width=0.35\textwidth]{Globecom2024_Draft/Figures/20.jpg}
%          \caption{Network playground.}
% \label{fig:network}
% \end{figure}
%
\begin{figure}[t]
      \centering           \includegraphics[width=0.48\textwidth]{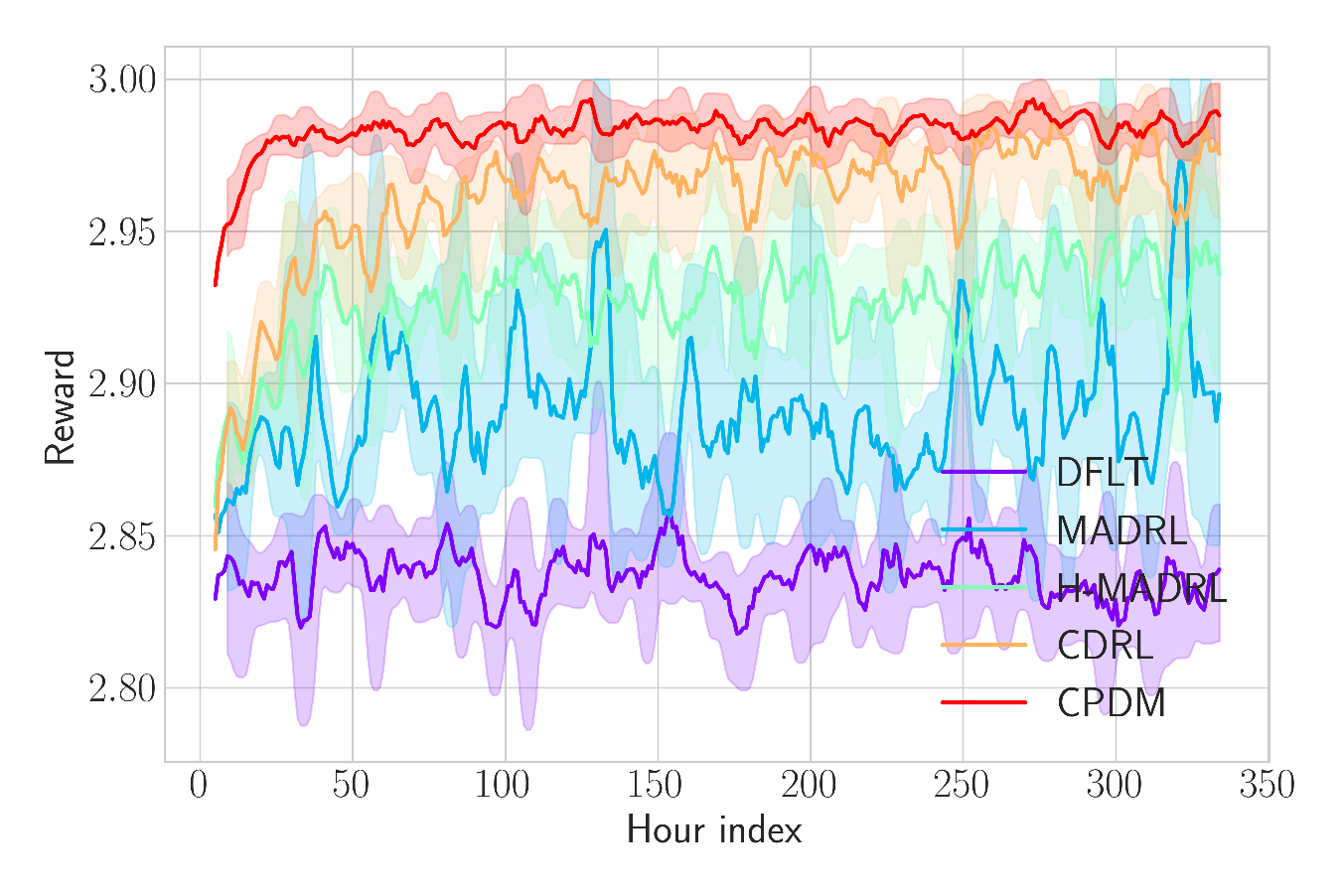}
          \caption{Performance comparison: Reward evolution during training.}
 \label{fig:rewards}
 \end{figure}
\subsubsection{Methods to Comparison}
We compare the performance of the following five methods:
\begin{itemize}
\item {\bf DFLT}: Default setting where all \acp{TTT} values are set to $320$ ms and \acp{CIO} values are set to $0$ dB.
\item {\bf MADRL}: The conventional baseline neighbor-aware per-cell agent scheme using \ac{DRL} as introduced in Table  \ref{tab:agent_compare}.
\item {\bf H-MADRL}: \ac{MADRL} with the two-tier heterogeneous agents introduced in Section \ref{ssec:localagents}, but without compositional learning.
\item {\bf CDRL}: \ac{MADRL} with both two-tier heterogeneous agents and compositional learning proposed in Section \ref{ssec:CDRL}.
\item  {\bf CPDM}: The two-tier agent scheme using compositional predictive optimization, proposed in Section \ref{ssec:CPDM}.
\end{itemize}

\subsubsection{Training Performance}
In Figure \ref{fig:rewards}, we compare the reward evolution of the five methods discussed. It is evident that MADRL struggles to converge within the constrained training timeframe. H-MADRL shows a slight improvement over MADRL by leveraging its two-tier modular light models. When both the two-tier design of heterogeneous agents and compositional learning are applied, \ac{CDRL} achieves a further acceleration in the convergence speed.

However, due to the limited sample size, the DRL-based approaches still fail to converge within the given training period. This is likely attributed to the joint training of actor and critic networks, which can introduce instability and demand longer training durations for effective convergence. In contrast, \ac{CPDM} significantly outperforms all DRL-based methods. Its superior performance stems from the efficient learning facilitated by lightweight prediction models and robust prediction-based decision-making.

\subsubsection{Inference Performance}
Table \ref{tab:KPI_compare} presents a comparison of the averaged \acp{KPI} over the evaluation period. Throughput and \ac{RLF} anomalies are identified by analyzing performance deviations from the default configuration for each hour of the day and each cell. Consequently, these values are marked as N/A for DFLT. The baseline cell-based MADRL approach shows unsatisfactory performance, likely due to its complexity, which hinders full convergence. It often over-corrects too-early \ac{HO} behaviors, leading to an increase in \ac{RLF} and HOL events. Note that the lowest latency observed for MADRL may result from excessive user dropping.

\begin{figure}[t]
     \centering
     \begin{subfigure}{.48\textwidth}
         \centering         \includegraphics[width=\textwidth]{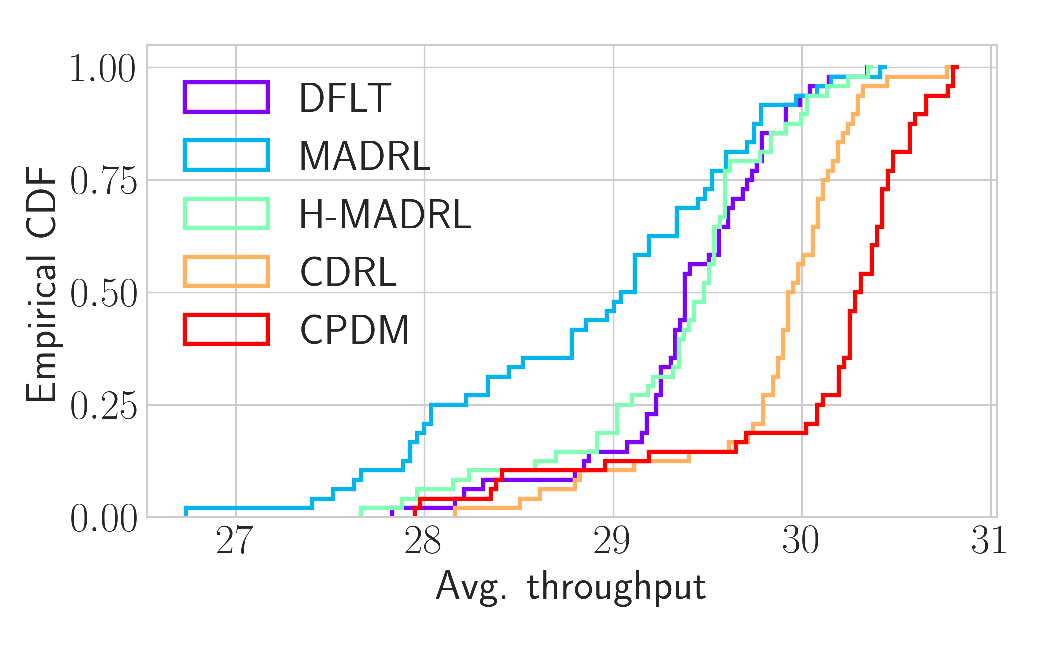}
         \caption{CDF of average throughput.}
         \label{fig:cdf_thr}
     \end{subfigure}
     \begin{subfigure}{0.48\textwidth}
         \centering
    \includegraphics[width=\textwidth]{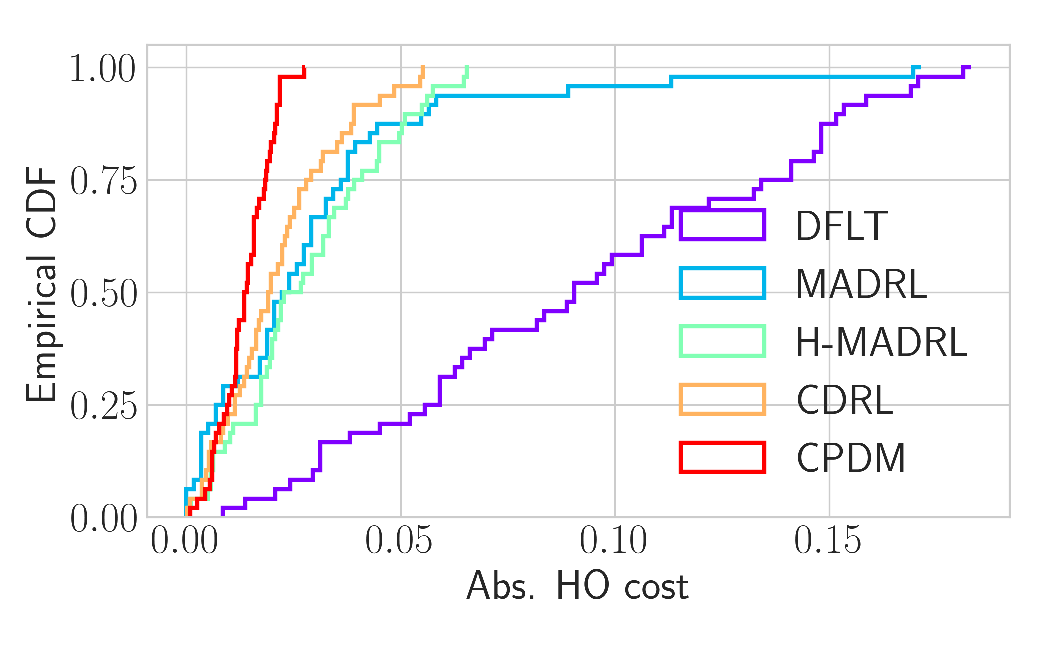}
         \caption{CDF of absolute HO cost.}
         \label{fig:cdf_hoc}
     \end{subfigure}
        \caption{Performance comparison of CDFs.}
        \label{fig:perf_comp_cdf}
\end{figure}

Figure \ref{fig:perf_comp_cdf} depicts the empirical \acp{CDF} of average throughput and absolute \ac{HO} cost. Notably, a \ac{HO} cost approaching $0$ reflects an effective trade-off in \ac{HO} behavior, balancing between too-late and too-early \acp{HO}.

Ultimately, compared to the default configuration, CPDM achieves a $33.3\%$ reduction of too-late \acp{HO} and $30.2\%$ reduction of too-early \acp{HO}, with slight improvement in throughput and latency. Compared with CDRL, CPDM significantly reduces the throughput and \ac{RLF} anomalies.

\subsubsection{Accuracy of Compositional Predictive Functions}
 The compositional learning of the prediction functions achieves accurate prediction results, with $0.006$ mean absolute error on HO cost, $86.8\%$ precision and  $81.1\%$ recall on throughput classification, and $94.6\%$ precision and $92.0\%$ recall on \ac{RLF} anomaly detection (plots omitted due to the limited space).    

\subsubsection{Key Takeaways}
We summarized the takeaways from the numerical results as follows:
\begin{itemize}
\item Due to practical limitations in learning time, the MADRL baseline approach fails to achieve convergence. The proposed two-tier agent design, featuring cooperating CAs and CPAs, effectively improves learning speed. In addition, compositional learning of the reward function further accelerates convergence.
\item  Under limited sample availability, predictive decision-making approaches exhibit greater stability and effectiveness compared to DRL-based methods. The proposed two-tier heterogeneous agent framework, combined with compositional learning, significantly reduces model complexity, enhances collaborative capabilities, and improves model reusability across different SON objectives.  
\end{itemize}

\section{Conclusion}\label{sec:Conclusion}
This paper studies multi-agent \ac{SON} network systems, seeking to alleviate complexities and bolster modularity through the exploration of two compositional learning methods: \ac{CDRL} and \ac{CPDM}. By leveraging a modular framework comprising two-tier distributed agents, we effectively mitigate model complexity while enhancing knowledge transferability. The numerical assessments demonstrate the remarkable efficacy of our approach, particularly in terms of sample efficiency, convergence rate, and training safety. These findings suggest that prediction-based decision-making methods may present a more stable and effective performance than the \ac{DRL} methods in industrial applications with costly training periods.

\bibliographystyle{IEEEtran}
\bibliography{references}

%\appendix
%\input{Content/appendix}

\end{document}